\documentclass[conference]{IEEEtran}
\IEEEoverridecommandlockouts
\usepackage{cite}
\usepackage{amsmath,amssymb,amsfonts}
\usepackage{algorithmic}
\usepackage{graphicx}
\usepackage{textcomp}
\usepackage{multirow}
\usepackage{xcolor}
\usepackage{algorithm}
\usepackage{algorithmic}
\usepackage{booktabs}
\def\BibTeX{{\rm B\kern-.05em{\sc i\kern-.025em b}\kern-.08em
    T\kern-.1667em\lower.7ex\hbox{E}\kern-.125emX}}
    \usepackage[left=0.68in,right=0.68in,top=0.75in,bottom=1.1in]{geometry}
\begin{document}

\title{Towards Imbalanced Large Scale Multi-label Classification with Partially Annotated Labels
}

\author{\IEEEauthorblockN{1\textsuperscript{st} Xin Zhang}
\IEEEauthorblockA{
\textit{University of South Carolina}\\
Columbia, United States \\
xz8@email.sc.edu}
\and
\IEEEauthorblockN{2\textsuperscript{nd} Yuqi Song}
\IEEEauthorblockA{
\textit{University of South Carolina}\\
Columbia, United States \\
yuqis@email.sc.edu}
\and
\IEEEauthorblockN{3\textsuperscript{rd} Fei Zuo}
\IEEEauthorblockA{
\textit{University of Central Oklahoma}\\
Oklahoma, United States \\
fzuo@uco.edu}
\and

\IEEEauthorblockN{4\textsuperscript{th} Xiaofeng Wang}
\IEEEauthorblockA{
\textit{University of South Carolina}\\
Columbia, United States \\
wangxi@cec.sc.edu}
}
\maketitle

\begin{abstract}
Multi-label classification is a widely encountered problem in daily life, where an instance can be associated with multiple classes. In theory, this is a supervised learning method that requires a large amount of labeling. However, annotating data is time-consuming and may be infeasible for huge labeling spaces. In addition, label imbalance can limit the performance of multi-label classifiers, especially when some labels are missing. Therefore, it is meaningful to study how to train neural networks using partial labels.
In this work, we address the issue of label imbalance and investigate how to train classifiers using partial labels in large labeling spaces. First, we introduce the pseudo-labeling technique, which allows commonly adopted networks to be applied in partially labeled settings without the need for additional complex structures. Then, we propose a novel loss function that leverages statistical information from existing datasets to effectively alleviate the label imbalance problem. In addition, we design a dynamic training scheme to reduce the dimension of the labeling space and further mitigate the imbalance.
Finally, we conduct extensive experiments on some publicly available multi-label datasets such as COCO, NUS-WIDE, CUB, and Open Images to demonstrate the effectiveness of the proposed approach. The results show that our approach outperforms several state-of-the-art methods, and surprisingly, in some partial labeling settings, our approach even exceeds the methods trained with full labels.

\end{abstract}


\section{Introduction}
The task of classifying a set of instances, each associated with a unique class label from a set of disjoint class labels, is known as multi-class (MC) classification. 
MC classification is the most common task in the early stage of the development of deep learning. 
However, as the importance of the role assumed by deep learning increases, multi-label (ML) classification, a generalized version of the MC problem, attracts more attention: due to the intrinsic plurality of the physical world~\cite{durand2019learning}, the multi-label classification is a more practical problem in our real lives, which allows the instances to be associated with more than one class~\cite{tsoumakas2007multi}.
For example, a CT scan can detect a variety of possible lesions to help people detect potential diseases early; materials have multiple properties such as pyroelectricity and hardness~\cite{song2021computational}; and for vision-based autonomous driving, the ego-vehicle needs to examine surroundings to find which objects are contained in the current scene and take corresponding actions~\cite{zhang2022effective}. 

In theory, ML classification is a form of supervised learning that necessitates a large number of accurate labels~\cite{hastie2009overview}.
However, in practice, annotating all labels for each training instance raises a great challenge in multi-label classification, which is time-consuming and even impractical, especially in the presence of a large number of categories~\cite{cole2021multi,zhang2021simple}. 
Therefore, training a classifier with the partially annotated labels (PAL) setting arouses extensive research interest.
In this setting, for each instance, only a subset of all the labels are annotated and the rest remains unknown. The acquisition of datasets in the PAL setting is much simpler and saves more annotation efforts compared with the scenario of the fully annotated labels (FAL) setting. 
There are two variations of the PAL setting: the partially annotated positive labels (PPL) and the single annotated positive label (SPL).
In the former case, all negative labels are discarded and only a portion of positive labels are retained. However, in the latter case, there is only one positive label and the others are missing, which is more extreme.
Table~\ref{table1} shows the difference between these settings.
\begin{table}[!tbp]
\caption{Different Missing-label Settings. \checkmark, $\times$, $?$ represent that the current instance belongs to this class, does not belong to this class, and lacks related labels, respectively.}
\begin{center}
\begin{tabular}{|c|c|c|c|c|c|}
\hline
\textbf{Settings}&\textbf{\textit{Class 1}}&\textbf{\textit{Class 2}}&\textbf{\textit{Class 3}}&\textbf{\textit{Class 4}}&\textbf{\textit{Class 5}} \\
\hline
FAL&\checkmark&$\times$&\checkmark &$\times$ &\checkmark  \\
\hline
PAL&\checkmark&$?$&$?$ &$\times$ &\checkmark  \\
\hline
PPL&\checkmark&$?$&\checkmark &$?$ &$?$  \\
\hline
SPL&\checkmark&$?$&$?$ &$?$ &$?$  \\
\hline
\end{tabular}
\label{tab1}
\end{center}
\label{table1}
\end{table}

Label imbalance is an issue that requires extra attention for multi-label classification problems~\cite{lin2017focal,zhang2022effective}.
Even in the FAL setting, there is a significant imbalance for many commonly used datasets.
As an example, the COCO dataset~\cite{coco} has only 2.9 positive labels out of 80 labels per image, and in the NUS-WIDE dataset~\cite{nus}, each image has an average of only 1.9 positive labels and 78.1 negative labels. 
This issue is even severer for partially labeled settings, where the few existing positive labels may also be discarded (an image may correspond to \textbf{several} positive labels and \textbf{zero} negative label), leading the classifier to generate trivial solutions where it simply predicts all categories as negative.
This problem significantly limits the performance of classifiers and the research on the partially labeled classification. 
Most existing approaches require that there is at least one positive label per image, i.e., PPL or SPL setting instead of confronting these challenges directly~\cite{cole2021multi,zhang2021simple}.
In addition, scaling up to the problems with large labeling space has not been sufficiently addressed\cite{yu2014large}.

In this paper, we focus on the issue of label imbalance and investigate how to address it in partially labeled settings. Also, how to train classifiers using partial labels with large labeling space will be taken into consideration. 
Specifically, We divide the labels of each instance in the training set into two parts, i.e., \textit{label existing} and \textit{label non-existing}.
For the \textit{non-existing} part, we introduce pseudo-labels as the target when calculating the loss.
At the same time, we directly take the labels as the target for the \textit{existing} part. In this way, the conventional neural-networks-based classifier can be directly applied to the partially labeled settings without adding other auxiliary structures which usually introduce extra burdens for the computation and training. 
Then, we take full advantage of the existing statistical information of the dataset and design a novel dynamic loss function that balances the model's attention to both \textit{existing} and \textit{non-existing} parts during the training.
In addition, we propose a new training scheme to further alleviate the label imbalance problem. 
During the training, we dynamically adjust the size of the \textit{non-existing} part by random sampling and continuously modify the learning rate as training progresses to ensure that the model does not produce trivial solutions due to a sudden influx of a large number of negative labels.


The main contributions are summarized as follows:
\begin{itemize}
    \item We propose a method for training multi-label classification models using partially labeled data based on the pseudo-labeling technique. This method can effectively reduce the manual labeling effort while ensuring the accuracy of the model.
    \item We address the issue of label imbalance in multi-label problems, even in large-scale labeling spaces, by designing sophisticated loss functions that utilize the statistical information of existing datasets and dynamically adjusting the training scheme.
    \item The extensive experiments on four large-scale public image datasets (COCO, NUS-WIDE, CUB, Open Images) demonstrate that our method outperforms the state-of-the-art methods.
\end{itemize}

The rest of the paper is organized as follows.  Section~\ref{sec:related} discusses the related work.  The problem is formulated in Section~\ref{sec:pre} and our proposed method is presented in Section~\ref{sec:PM}.  Section~\ref{sec:exp} shows the experimental results.  Finally, conclusions are drawn in Section~\ref{sec:con}.

\section{Related Work}
\label{sec:related}
From the early stages of computer vision, multi-label problems have been a hot topic in the research community. Therefore, in this section, we will only introduce the work that is most relevant to our research  due to space limitations.

\subsection{Partial Labels}
Due to the extremely expensive manual labeling costs, ML tasks often involve incomplete training data~\cite{durand2019learning}.
By treating the missing labels as negative labels, the partially labeled setting then becomes a fully labeled learning problem~\cite{bucak2011multi,mahajan2018exploring,chen2013fast}.
However, performance drops because a lot of ground-truth positive labels are initialized as negative labels~\cite{joulin2016learning}, i.e., labels imbalance.
Another solution is to consider each label prediction as an independent binary classification problem~\cite{tsoumakas2007multi}.
But this approach is not scalable when the number of categories grows and it ignores the correlations amongst labels and the relations amongst instances~\cite{durand2019learning}.
In addition, there are several works on propagating the information of existing labels to the missing parts by utilizing label correlation techniques such as matrix completion algorithm~\cite{cabral2011matrix,xu2013speedup}, low-rank empirical risk minimization~\cite{yu2014large}, and mixed graph~\cite{wu2015ml}.
However, most of these works require solving an optimization problem with the training set in memory, so it is impossible to use a mini-batch strategy to fine-tune the model~\cite{durand2019learning}.

\subsection{Label Imbalance}
Label imbalance is a huge challenge for lots of computer vision-related tasks, such as multi-label classification and object detection~\cite{ridnik2021asymmetric}. 
The issue is stated as there is only a small fraction of possible labels for most images, which means that the number of positive labels is much lower than the number of negatives.
Several methods are presented to address it.
\cite{wu2020distribution} proposes a loss function to alleviate the imbalance, but this loss function is only aimed at long-tail distribution scenarios.
Focal loss function~\cite{lin2017focal}, which is presented to solve the imbalance in object detection first, can also be used for ML classification problems.
However, Focal treats the positive and negative samples equally and it will result in the accumulation of more loss gradients from negative samples and down-weighting of important contributions from the rate positive samples~\cite{ridnik2021asymmetric}. 
Based on resampling methods, \cite{oksuz2020imbalance} selects only a subset of the possible examples to solve the problem of imbalance in object detection.
But resampling methods are not suitable for multi-label classification, since they cannot change the distribution of only a specific label~\cite{ridnik2021asymmetric}.
In the setting of PAL, this problem is more severe because the number of positive labels may further decrease, and the phenomenon of imbalance will become more obvious~\cite{zhang2022effective}.
Therefore, studying how to solve imbalance is of great significance.

\subsection{Large Labeling Space}
The focus of recent research on ML classification has largely been shifted to the issue that the number of labels is assumed to be extremely large, with the key challenge being the design of scalable algorithms that offer real-time predictions and have a small memory consumption~\cite{yu2014large}.  
Common solutions are proposed that either reduce the dimension of the labels, such as the Compressed Sensing Approach~\cite{hsu2009multi}, CPLST~\cite{chen2012feature}, and Bayesian CS~\cite{kapoor2012multilabel}, or reduce the feature dimension~\cite{sun2010canonical}, or both, such as WSABIE~\cite{weston2010large}.
However, these previous work cannot handle partial labeling settings and is tied to a specific loss function~\cite{yu2014large}.

\section{Preliminary }
\label{sec:pre}

\subsection{Multi-label Classification}
Given a multi-label classifier with full labels, let $\mathcal{X}=\mathbb{R}^M$ be the input attribute space of $M$-dimensional feature vectors and $\mathcal{Y}=\{1, 2,\ldots,L\}$ denotes the set of $L$ possible labels. 
An instance $\textbf{x}\in \mathcal{X}$ is associated with a subset of labels $\textbf{y}\in2^\mathcal{Y}$, which can be represented as an $L$-vector $\textbf{y}=[y_1,y_2,\ldots,y_L]=\{0,1\}^L$ where $y_j=1$ if and only if the $j$th label is relevant (otherwise $y_j=0$).
Let $\mathcal{D}=\{(\textbf{x}_1, \textbf{y}_1),\ldots,(\textbf{x}_N, \textbf{y}_N) \}$ is the training set of $N$ samples.
Given $\mathcal{D}$, a multi-label classifier $h:\mathcal{X}\rightarrow\mathcal{Y}$ learns to map the attribute input space to the label output space. We use $\hat{\textbf{y}}$ to present the prediction of classifier $h$, that is, $\hat{\textbf{y}}=f(h(\textbf{x}))$, where $f(\cdot)$ stands for a function (commonly the sigmoid function as $\sigma(s)=\frac{1}{1+e^{-s}}$) that turns confidence outputs into a prediction. In this case, most of the existing work \cite{nam2014large,tsoumakas2007multi,wang2016cnn,sorower2010literature} adopts the binary cross entropy (BCE) function as the loss function, which is formulated by
\begin{equation}
    \mathcal{L}(\hat{y},y)=-\frac{1}{L}\sum_{i=1}^{L}[(y_i \log(\hat{y}_i)+(1-y_i)\log(1-\hat{y}_i)]\label{bce}.
\end{equation}

For multi-label classification with missing-label, we use $\textbf{z}=[z_1,z_2,\ldots,z_L]=\{0,1,\varnothing \}^L$ to represent the observed labels, where $z_j=\varnothing$ means the corresponding label is missing. 
It is worth mentioning that the training set is $\mathcal{D}=\{(\textbf{x}_1, \textbf{z}_1),\ldots,(\textbf{x}_N, \textbf{z}_N) \}$, while validation and test set is $\mathcal{D}_v=\mathcal{D}_t=\{(\textbf{x}_1, \textbf{y}_1),\ldots,(\textbf{x}_N, \textbf{y}_N) \}$ in this case. In other words, we still use full labels for validation and testing.

\subsection{Different Missing-label Settings}
According to the number and positive/negative properties of the observable labels, we divide the multi-label classification problem into several settings:
partially annotated labels (PAL), partially annotated positive labels (PPL), and an extreme case, i.e., single annotated positive label (SPL) \cite{cole2021multi}. Specially, we formulate these settings as the following:
\begin{equation}
\begin{aligned}
    &\textbf{z}_{PAL}=\{0,1,\varnothing\}^L \text{ and }\\ &\sum_{j=1}^L(\mathbf{1}_{[\textbf{z}_{PAL_j}=1]}+\mathbf{1}_{[\textbf{z}_{PAL_j}=0]})< L,
\end{aligned}
\end{equation}
\begin{equation}
\textbf{z}_{PPL}=\{1,\varnothing\}^L \text{ and } \sum_{j=1}^L\mathbf{1}_{[\textbf{z}_{PPL_j}=1]}< L,
\end{equation}
\begin{equation}
\textbf{z}_{SPL}=\{1,\varnothing\}^L \text{ and } \sum_{j=1}^L\mathbf{1}_{[\textbf{z}_{PPL_j}=1]}=1,
\end{equation}
where $\mathbf{1}_{[\cdot]}$ stands for the indicator function, that is, $\mathbf{1}_{[True]}=1$ and $\mathbf{1}_{[False]}=0$. Considering a large number of unknown labels and the imbalance problem in these settings, we design a new loss function to handle them, instead of directly using BCE~\eqref{bce}.

\section{Proposed Method}
\label{sec:PM}
\begin{figure*}[ht]
  \centering
   \includegraphics[width=0.95\textwidth]{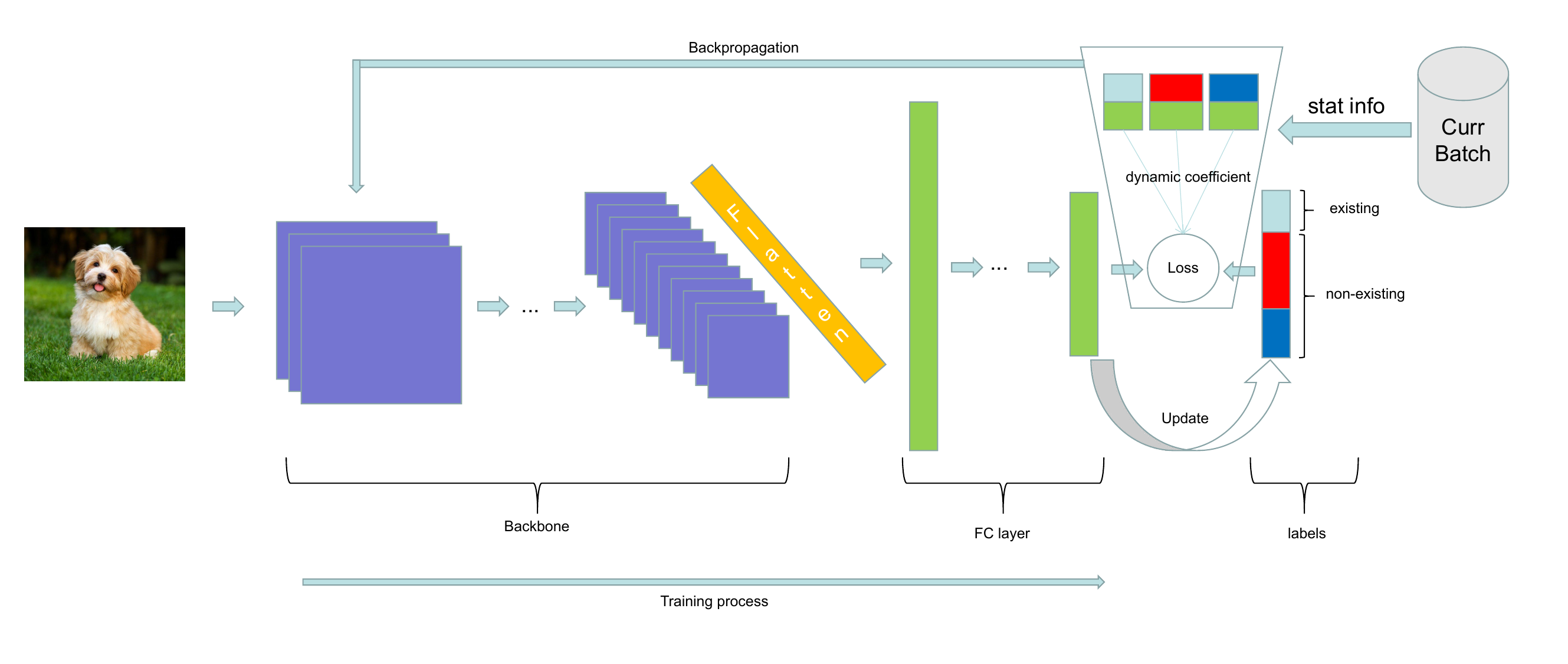}
    \caption{The pipeline of our proposed method. The direction of training progress is indicated from left to right. Our method does not rely on any additional structures, so the model structure is composed only of a backbone and FC layers. We input the statistical information of the current batch, the predicted results obtained from training, and the ground-truth labels that the current image has (existing parts, in light blue) and the pseudo labels (non-existing parts), which are further divided into the picked parts (in red) and unpicked parts (in dark blue), into the loss function we proposed for calculation. The dynamic parameters in the loss function are continuously adjusted as the training progresses. Then, the loss value is backpropagated to update the parameters in the model.}
	\label{fig:pipeline}
\end{figure*}

Overall, the pipeline of our proposed method is shown as Fig.~\ref{fig:pipeline}.
Next, we will introduce three important components: the pseudo labels technique, the novel loss function, and the dynamic training scheme.

\subsection{Pseudo Labeling}
For multi-label classification problems, the most commonly used solution is to employ Binary Cross-Entropy (BCE)~\ref{bce} as the loss function and train the model using supervised learning to minimize the BCE loss value.
However, in the case of partial labeling settings, as some labels are missing for each image, we cannot directly use BCE as the loss function.
To avoid introducing the extra overhead and thus reducing the efficiency of training and prediction (such as the need to add an additional GNN network~\cite{durand2019learning} or use another network with the same structure as the label estimator~\cite{cole2021multi}), we choose to introduce the pseudo-labeling technique to transform the partially labeled setting into a fully supervised learning method without changing the network structure.
Specifically, we divide the labels corresponding to the image into two categories: one category corresponds to the existing labels, and the other corresponds to the missing labels. 
In this way, by introducing pseudo labels, we can apply commonly used supervised learning methods to partially labeled settings without the need for major changes to the model structure.

The two key points of the pseudo-labeling technique are the initialization and updating of pseudo labels~\cite{zhang2022effective}. 
In our method, we simply initialize all pseudo labels to 0.5 and use the model's prediction results as the updated value for these pseudo labels.
We formulate these processes as~\eqref{initandup},
\begin{equation}
    \begin{aligned}
        \widetilde{y}_{i_j} = \begin{cases}
    \hat{y}_{i_j}, & \textit{update}\\
    0.5, & \textit{initialization}
    \end{cases}\label{initandup}
    \end{aligned}
\end{equation}
where $\widetilde{y}_{i_j}$ and $\hat{y}_{i_j}$ stands for the pseudo label and the prediction of model for $i$th image's $j$th label, respectively.



\subsection{Loss Function Designing}
Our loss function is also divided into two parts, just like we divided the labels corresponding to an image instance into the \textit{existing} and the \textit{non-existing} parts. 
The overall loss function in our approach is shown as~\eqref{eq:overall},
\begin{equation}
    \mathcal{L} = \mathcal{L}_{\text{Exist}} +  \mathcal{L}_{\text{Non-exist}}.\label{eq:overall}
\end{equation}
The simplest way to implement $\mathcal{L}_{\text{Exist}} $ and $\mathcal{L}_{\text{Non-exist}}$ is to use BCE function. 
That is,
\begin{equation}
\begin{aligned}
    \mathcal{L}_{\text{Exist}}&=\mathcal{L}_{BCE}(\hat{y},y),\\
    \mathcal{L}_{\text{Non-exist}}&=\mathcal{L}_{BCE}(\hat{y},\widetilde{y}).
    \end{aligned}
    \label{bcefull}
\end{equation}

However, this approach suffers from label imbalance, particularly when considering the case where positive labels for the current image may not exist, which may lead to trivial solutions.
Therefore, our loss function does not directly use BCE to implement it but adopts an imbalance-resistant design.
Specifically, we add a penalty coefficient to~\eqref{bcefull}, which drives the model to be more inclined to predict a positive label.
After adding the penalty coefficient and merging these two parts together, we obtained the loss function~\eqref{bcewithpen},
\begin{equation}
\begin{aligned}
    \mathcal{L} =& -\alpha[\frac{1}{N_{\text{e}}}\sum_{i=1}^{N_{\text{e}}}(y_i \log(\hat{y}_i)) + \frac{1}{N_{\text{ne}}}\sum_{i=1}^{N_{\text{ne}}}(\widetilde{y}_i \log(\hat{y}_i))]\\
    &-\beta[\frac{1}{N_{\text{e}}}\sum_{i=1}^{N_{\text{e}}}((1-y_i) \log(1-\hat{y}_i)) \\
    &+ \frac{1}{N_{\text{ne}}}\sum_{i=1}^{N_{\text{ne}}}((1-\widetilde{y}_i) \log(1-\hat{y}_i))]
\end{aligned}
    \label{bcewithpen}
\end{equation}
where $\alpha$ and $\beta$ are penalty coefficients, $N_{\text{e}}$ and $N_{\text{ne}}$ stand for the number of existing labels and pseudo labels for the current image.
During training, we work on a batch level rather than the entire dataset, so the statistical information of the batch is more refined and more instructive than that of the entire dataset.
Therefore, in our approach, the value of the penalty coefficient is obtained from the label proportion information of the current batch, and the values of coefficients are different for different batches.
Specifically, we set $\beta=\max(\frac{P}{T},\gamma)$ and $\alpha = 1-\beta$, where $\gamma$ is a hypermeter to avoid the value of $\alpha$ is so large that it dominates the loss calculation. 
We set $\gamma=0.05$, which is determined by extensive experiments.

In addition, considering that the initial values of the pseudo labels are 0.5, we can not treat the loss of the existing and non-exist parts equally in the early stages of training. 
Our goal is for the model to prioritize the existing part in the early stages of training and gradually shift its focus towards the non-existing part.
Therefore, we introduce a dynamic coefficient $d$, which is determined by the current epoch index, as~\eqref{d},
\begin{equation}
  d=e^{\frac{e_c+m}{e_T}},  \label{d}
\end{equation}
where $e_c$ and $e_T$ stand for the current epoch index and the total epoch number respectively, and $m$ is a hyperparameter for the non-exist part, which gradually increases with the progress of training.
Obviously, when it is the $e_T-m$-th epoch, the model treats the existing and non-existing parts equally. If the current epoch is less than $e_T-m$, the model is more inclined towards the existing part, while if it is greater than $e_T-m$, the opposite is true.
Therefore, the larger the value of $m$, the faster the model will focus on the non-existent part. In our experiments, we set $m=3$ in the following.

In addition, to further alleviate the label imbalance, we also designed a penalty coefficient $p$ to current loss, as shown in~\eqref{penterm},
\begin{equation}
    p= 1+|sign(\sum_{i=1}^Nmax(\hat{y}_i-0.1,0))|                ,\label{penterm}
\end{equation}
where $sign()$ is the sign function.
When the prediction values for all classes are less than 0.1, $p=2$, otherwise, $p=1$.
Obviously, this penalty term will attempt to prevent the model from producing all negative predictions by doubling the current loss value for any input image. 
Finally, we obtain the loss function as~\eqref{final},
\begin{equation}
\begin{aligned}
    \mathcal{L} =& -p\alpha[\frac{1}{N_{\text{e}}}\sum_{i=1}^{N_{\text{e}}}(y_i \log(\hat{y}_i)) + \frac{d}{N_{\text{ne}}}\sum_{i=1}^{N_{\text{ne}}}(\widetilde{y}_i \log(\hat{y}_i))]\\
    &-p\beta[\frac{1}{N_{\text{e}}}\sum_{i=1}^{N_{\text{e}}}((1-y_i) \log(1-\hat{y}_i)) \\
    &+ \frac{d}{N_{\text{ne}}}\sum_{i=1}^{N_{\text{ne}}}((1-\widetilde{y}_i) \log(1-\hat{y}_i))]
\end{aligned}
\label{final}
\end{equation}



\subsection{Dynamic Training Scheme}
In our approach, we further suppress class imbalance and address the large labeling space problem by modifying the training scheme.
After dividing the corresponding labels of each image into existing and non-existing parts, we further split the non-existing part.
Specifically, we randomly select $q$\% of the labels in the non-existing part to participate in training, ignoring the rest.
In the training process, our loss function is actually as shown in~\eqref{dynamicloss},
\begin{equation}
    \begin{aligned}
    \mathcal{L} =& -p\alpha[\frac{1}{N_{\text{e}}}\sum_{i=1}^{N_{\text{e}}}(y_i \log(\hat{y}_i)) + \frac{d}{q N_{\text{ne}}}\sum_{i=1}^{q N_{\text{ne}}}(\widetilde{y}_i \log(\hat{y}_i))]\\
    &-p\beta[\frac{1}{N_{\text{e}}}\sum_{i=1}^{N_{\text{e}}}((1-y_i) \log(1-\hat{y}_i)) \\
    &+ \frac{d}{q N_{\text{ne}}}\sum_{i=1}^{q N_{\text{ne}}}((1-\widetilde{y}_i) \log(1-\hat{y}_i))]\\
    &+ \mathcal{L}_{\text{ned}},
\end{aligned}
\label{dynamicloss}
\end{equation}
where $\mathcal{L}_{\text{ned}}$ stands for the discarded part of non-existing, and it always equals 0. 
It should be pointed out that our model still predicts all categories, but only q\% of the non-existing part is involved in the loss calculation.
This means that we control the proportion of partially labeled data during training, and increase the proportion of existing labels in the total training labels. This makes the model's prediction results more reliable.

BR is a related approach that breaks down the multi-label problem into several binary classification problems, but its performance is inadequate as it fails to account for the interdependence between labels, particularly in the context of large labeling spaces. However, although our method also involves dividing the labels, for each image we randomly partition the labels, and theoretically, the selected labels for each training epoch are different. Therefore, the randomness ensures that our method can retain the dependence between labels while reducing the label space.

\section{Experiments}
\label{sec:exp}
In this section, we first provide an introduction to the selected datasets and the data pre-processing method to obtain their partially labeled versions.
Then, we describe the metrics and baseline methods used to evaluate our approach. 
Finally, we analyze the experimental results.

\subsection{Datasets}
We perform experiments on several standard multi-label datasets: COCO~\cite{coco}, NUS-WIDE~\cite{nus}, CUB~\cite{cub}, and Open Images~\cite{kuznetsova2020open}.
For these four datasets, we split the training and testing sets as the official splitting method. 
The statistical information for these datasets is summarized in Table~\ref{table:ds}. 
Specifically, the COCO, NUS, and CUB datasets are fully labeled. For COCO, each image has an average of 2.9 positive labels and 77.1 negative labels. For NUS and CUB, each image has an average of 1.9 positive labels and 78.1 negative labels, and 31.3 positive labels and 280.7 negative labels, respectively.
In contrast, Open Images only has partially labeled data, with each image corresponding to 600 labels. However, on average, only 5 labels per image are present.

\begin{table}[htbp]
\begin{center}

\caption{The statistical information of the four datasets.}
\begin{tabular}{|l|l|l|l|l|}
\hline
                   & COCO & NUS & CUB & Open Images \\ \hline
The size of training set      &  82,081    &  126,034   &  5,994   &     5,655,108        \\ \hline
The size of test set          &   40,137   &  84,226   &   5,794  &     125,436        \\ \hline
Positive per image &  2.9    &  1.9   &    31.3 &       2.4      \\ \hline
Negative per image &   77.1   &  78.1   &   280.7  &     2.6        \\ \hline
Total categories   &   80   &   80  &   312  &     600        \\ \hline
\end{tabular}
\label{table:ds}
\end{center}
\end{table}

In order to obtain the partially labeled variants of COCO, NUS-WIDE and CUB, inspired by~\cite{cole2021multi}, 
We randomly select a certain proportion of labels to be retained and discard the rest. 
Specifically, for SPL, we keep only one positive label and discard all other labels. 
For PPL, we retain a certain proportion of positive labels and discard the remaining negative labels.
And for PAL, we randomly keep a certain proportion of labels, regardless of whether they are positive or negative. 
Therefore, in PAL setting, an image may correspond to several negative labels but zero positive labels. 
It should be noted that for these datasets, we only perform the above pre-processing methods once before training, and will not repeat them in subsequent training to ensure that the randomness of label selection will not affect the reliability of the experimental results.

As for the data augmentation, we first resize the original input image of all these three datasets to the shape of 448 × 448. And then, the horizontal flip is applied with a probability of 0.5. In the end, by using the standard ImageNet statistics, we normalize the input images.

\subsection{Baselines and Metrics}
In order to prove the effectiveness of our method, we first choose \textbf{BCE} and \textbf{BCE-LS} in full labeling settings as baselines. 
These two methods are the most commonly used methods in multi-label problems. The former takes BCE~\eqref{bce} as the loss function and the latter uses label smoothing BCE~\cite{szegedy2016rethinking}, which is proposed to reduce overfitting and has been shown to be effective in mitigating the negative impacts of label noises.

In addition, we also compare our method with \textbf{AN}~\cite{kundu2020exploiting}, \textbf{WAN}~\cite{mac2019presence},  \textbf{ROLE}~\cite{cole2021multi},  \textbf{Focal}~\cite{lin2017focal}, and \textbf{ASL}~\cite{ridnik2021asymmetric}. 
These methods are implemented slightly differently for different settings. 
For \textbf{AN}, \textbf{WAN}, and \textbf{ROLE}, we directly use the authors' source code and training methods without any modification in the PPL setting. 
But as mentioned in~\cite{zhang2022effective}, these algorithms require that each image has at least one available positive label. 
Therefore, before applying these methods to the PAL setting, we removed the restriction on the number of positive labels in the source code while keeping the rest unchanged. 
As for \textbf{Focal} and \textbf{ASL}, these two methods are originally designed to solve the label imbalance problem and cannot handle partially labeled problems. Therefore, to apply them to these partial settings and compare them with our method, we follow~\cite{bucak2011multi, mac2019presence,sun2010multi} to treat missing labels as negative labels.

As for the metrics, following~\cite{durand2019learning,cole2021multi,zhang2022effective}, we also utilize the mean average precision (mAP) to evaluate our approach.

\subsection{Network Structure}
We use an end-to-end network for all experiments: a ResNet-50 \cite{xie2017aggregated}, pre-trained on ImageNet \cite{deng2009imagenet}, as the backbone and fully connected layers, which is the same as the multi-label classifier under FOL setting. Our approach does not add any extra structure to the network.

\subsection{Experimental Results}
We first compare our approach with other baseline methods in four different settings on COCO, NUS-WIDE and CUB.

\smallskip
\noindent
\textbf{FAL.}
In FAL, all the labels are available.  
The results are summarized in Table~\ref{table.fal}.
It can be seen that our method outperforms other methods on all three datasets. Compared to the commonly used BCE loss function, our method improves performance by 4.3\%, 3.6\%, and 3.8\% on COCO, NUS-WIDE , and CUB, respectively. 
In comparison to Focal and ASL loss functions designed for dealing with labels imbalance, our method also shows significant improvements on all three datasets.

\begin{table}[htbp]
\begin{center}
\caption{The mAP results of our method and other baseline methods in FAL setting. We bold the best performance and underline the second best.}
\begin{tabular}{|c|c|c|c|}
\hline
       & ~~COCO~~~ & NUS-WIDE & ~~~CUB~~~ \\ \hline
BCE    &  75.8   &  52.6       & 32.1    \\ \hline
BCE-LS &  76.8    &   53.5       &   32.6  \\ \hline
Focal  &  79.6    &   55.3       & \underline{34.8}    \\ \hline
ASL    &  \underline{80.1}    &   \underline{55.9}       &  34.7   \\ \hline
Ours   &   \textbf{81.1}   &   \textbf{56.2}       &  \textbf{35.9}   \\ \hline
\end{tabular}
\label{table.fal}
\end{center}
\end{table}

\smallskip
\noindent
\textbf{PAL.}
For the three datasets, we keep 30\%, 60\%, and 90\% of the labels, and we refer to these settings as PAL\_0.3, PAL\_0.6, and PAL\_0.9, respectively.
The experimental results are shown in Table~\ref{table:pal}.
Our method achieves the best mAP scores in all settings. 
On the COCO dataset, our method achieves mAP scores of 65.6, 76.3, and 81.0 when retaining 30\%, 60\%, and 90\% of the labels, respectively. 
On the NUS dataset, the mAP scores are 42.3, 49.7, and 56.0, respectively. 
On the CUB dataset, the scores are 24.7, 32.2, and 36.0. 
Compared with the second-best algorithm, our method shows a significant improvement.
Surprisingly, on the COCO and CUB datasets, we exceed the performance of the BCE algorithm using only 60\% of the labels. 
This indicates that our method can save at least 40\% of manual annotation costs.
As AN, WAN, and ROLE algorithms require at least one positive label per image, their performance is significantly degraded in the PAL settings. 
This well demonstrates the effectiveness of our algorithm.

\begin{table*}[htbp]
\begin{center}
\caption{The mAP results in PAL settings. PAL\_0.3, PAL\_0.6, and PAL\_0.9 stands for the missing ratio of labels is 70\%, 40\% and 10\% respectively. We bold the best performance and underline the second best except BCE and BCE-LS there two strong baselines in these 9 different settings.}
\begin{tabular}{|c|ccc|ccc|ccc|}
\hline
                             & \multicolumn{3}{c|}{COCO}                                                & \multicolumn{3}{c|}{NUS-WIDE}                                            & \multicolumn{3}{c|}{CUB}                                                 \\ \hline
\multicolumn{1}{|c|}{BCE}    & \multicolumn{3}{c|}{75.8}                                                    & \multicolumn{3}{c|}{52.6}                                                    & \multicolumn{3}{c|}{32.1}                                                    \\ \hline
\multicolumn{1}{|c|}{BCE-LS} & \multicolumn{3}{c|}{76.8}                                                    & \multicolumn{3}{c|}{53.5}                                                    & \multicolumn{3}{c|}{32.6}                                                    \\ \hline
                             & \multicolumn{1}{c|}{PAL\_0.3} & \multicolumn{1}{c|}{PAL\_0.6} & PAL\_0.9 & \multicolumn{1}{c|}{PAL\_0.3} & \multicolumn{1}{c|}{PAL\_0.6} & PAL\_0.9 & \multicolumn{1}{c|}{PAL\_0.3} & \multicolumn{1}{c|}{PAL\_0.6} & PAL\_0.9 \\ \hline
AN                           & \multicolumn{1}{c|}{50.1}         & \multicolumn{1}{c|}{65.6}         &   72.7       & \multicolumn{1}{c|}{29.4}         & \multicolumn{1}{c|}{40.5}         &    50.1      & \multicolumn{1}{c|}{13.2}         & \multicolumn{1}{c|}{19.2}         &    27.4      \\ \hline
WAN                          & \multicolumn{1}{c|}{54.7}         & \multicolumn{1}{c|}{69.9}         & 77.0         & \multicolumn{1}{c|}{31.2}         & \multicolumn{1}{c|}{44.8}         & 52.6         & \multicolumn{1}{c|}{18.1}         & \multicolumn{1}{c|}{25.9}         &   30.3       \\ \hline
ROLE                         & \multicolumn{1}{c|}{53.6}         & \multicolumn{1}{c|}{68.1}         &  77.2        & \multicolumn{1}{c|}{30.9}         & \multicolumn{1}{c|}{45.1}         &    52.0      & \multicolumn{1}{c|}{17.6}         & \multicolumn{1}{c|}{25.4}         &  31.9        \\ \hline
Focal                        & \multicolumn{1}{c|}{\underline{62.1}}         & \multicolumn{1}{c|}{71.6}         &    78.8      & \multicolumn{1}{c|}{\underline{37.6}}         & \multicolumn{1}{c|}{45.9}         &    \underline{55.0}      & \multicolumn{1}{c|}{19.0}         & \multicolumn{1}{c|}{26.7}         &    \underline{34.2}      \\ \hline
ASL                          & \multicolumn{1}{c|}{60.9}         & \multicolumn{1}{c|}{\underline{72.8}}         &    \underline{79.9}      & \multicolumn{1}{c|}{35.9}         & \multicolumn{1}{c|}{\underline{47.0}}         &   54.3       & \multicolumn{1}{c|}{\underline{20.1}}         & \multicolumn{1}{c|}{\underline{29.6}}         &   34.0       \\ \hline
Ours                         & \multicolumn{1}{c|}{\textbf{65.6}}         & \multicolumn{1}{c|}{\textbf{76.3}}         &  \textbf{81.0}        & \multicolumn{1}{c|}{\textbf{42.3}}         & \multicolumn{1}{c|}{\textbf{49.7}}         &      \textbf{56.0}    & \multicolumn{1}{c|}{\textbf{24.7}}         & \multicolumn{1}{c|}{\textbf{32.2}}         &      \textbf{36.0}    \\ \hline
\end{tabular}
\label{table:pal}
\end{center}
\end{table*}

\smallskip
\noindent
\textbf{PPL.}
For the PPL setting, we only keep a certain percentage of positive labels and discard all negative labels. Specifically, we keep 30\%, 60\%, and 90\% of the positive labels for each dataset.
The experimental results are summarized in Table~\ref{table:ppl}.
Our method achieves the best mAP scores in all settings of PPL. Compared to the PAL setting, all methods show significant improvement, as the proportion of positive labels increases. In this setting, our method can outperform strong baseline algorithms such as BCE and BCE-LS with only 60\% of positive labels used.

\begin{table*}[htbp]
\begin{center}
\caption{The mAP results in PPL settings. PPL\_0.3, PPL\_0.6, and PPL\_0.9 stands for the missing ratio of positive labels is 70\%, 40\%, and 10\% respectively. We bold the best performance and underline the second best except BCE and BCE-LS there are two strong baselines in these 9 different settings.}
\begin{tabular}{|c|ccc|ccc|ccc|}
\hline
                             & \multicolumn{3}{c|}{COCO}                                                & \multicolumn{3}{c|}{NUS-WIDE}                                            & \multicolumn{3}{c|}{CUB}                                                 \\ \hline
\multicolumn{1}{|c|}{BCE}    & \multicolumn{3}{c|}{75.8}                                                    & \multicolumn{3}{c|}{52.6}                                                    & \multicolumn{3}{c|}{32.1}                                                    \\ \hline
\multicolumn{1}{|c|}{BCE-LS} & \multicolumn{3}{c|}{76.8}                                                    & \multicolumn{3}{c|}{53.5}                                                    & \multicolumn{3}{c|}{32.6}                                                    \\ \hline
                             & \multicolumn{1}{c|}{PPL\_0.3} & \multicolumn{1}{c|}{PPL\_0.6} & PPL\_0.9 & \multicolumn{1}{c|}{PPL\_0.3} & \multicolumn{1}{c|}{PPL\_0.6} & PPL\_0.9 & \multicolumn{1}{c|}{PPL\_0.3} & \multicolumn{1}{c|}{PPL\_0.6} & PPL\_0.9 \\ \hline
AN                           & \multicolumn{1}{c|}{63.2}         & \multicolumn{1}{c|}{67.0}         &    69.8      & \multicolumn{1}{c|}{44.2}         & \multicolumn{1}{c|}{48.3}         &     50.6     & \multicolumn{1}{c|}{18.6}         & \multicolumn{1}{c|}{26.3}         &   31.7       \\ \hline
WAN                          & \multicolumn{1}{c|}{68.1}         & \multicolumn{1}{c|}{70.8}         &     71.6     & \multicolumn{1}{c|}{46.3}         & \multicolumn{1}{c|}{48.5}         &  51.0        & \multicolumn{1}{c|}{21.6}         & \multicolumn{1}{c|}{28.9}         &      32.9    \\ \hline
ROLE                         & \multicolumn{1}{c|}{71.8}         & \multicolumn{1}{c|}{\underline{75.4}}         &  \underline{77.4}        & \multicolumn{1}{c|}{47.3}         & \multicolumn{1}{c|}{50.4}         &    52.5      & \multicolumn{1}{c|}{21.5}         & \multicolumn{1}{c|}{29.6}         &   33.1       \\ \hline
Focal                        & \multicolumn{1}{c|}{\underline{72.3}}         & \multicolumn{1}{c|}{75.0}         &  76.2        & \multicolumn{1}{c|}{\underline{48.1}}         & \multicolumn{1}{c|}{51.2}         &     52.1     & \multicolumn{1}{c|}{\underline{22.7}}         & \multicolumn{1}{c|}{\underline{30.2}}         &    \underline{34.6}      \\ \hline
ASL                          & \multicolumn{1}{c|}{71.9}         & \multicolumn{1}{c|}{\underline{75.4}}         & 76.8         & \multicolumn{1}{c|}{47.9}         & \multicolumn{1}{c|}{\underline{52.0}}         &   \underline{52.9}       & \multicolumn{1}{c|}{22.4}         & \multicolumn{1}{c|}{29.8}         &  34.5        \\ \hline
Ours                         & \multicolumn{1}{c|}{\textbf{73.1}}         & \multicolumn{1}{c|}{\textbf{80.1}}         &  \textbf{83.6}        & \multicolumn{1}{c|}{\textbf{50.2}}         & \multicolumn{1}{c|}{\textbf{54.7}}         & \textbf{59.9}         & \multicolumn{1}{c|}{\textbf{26.4}}         & \multicolumn{1}{c|}{\textbf{33.8}}         &  \textbf{37.9}        \\ \hline
\end{tabular}
\label{table:ppl}
\end{center}
\end{table*}

\smallskip
\noindent
\textbf{SPL.}
For the SPL setting, we only keep one positive label for each image and discard all other labels.
Table.~\ref{table.spl} summarizes the results.
It can be seen that our method has higher mAP scores than all other methods, especially for the COCO dataset, where our method is 6 points higher than the second-best method. However, because we only use one positive label, our method still lags behind the strong baseline even though it outperforms all the comparison methods.

\begin{table}[htbp]
\begin{center}
\caption{The mAP results in SPL setting. We bold the best performance and underline the second best except BCE and BCE-LS.}
\begin{tabular}{|c|c|c|c|}
\hline
       & ~~COCO~~~ & NUS-WIDE & ~~~CUB~~~ \\ \hline
BCE    &   75.8   &    52.6      &   32.1  \\ \hline
BCE-LS &  76.8    &   53.5       &   32.6  \\ \hline
AN     &    64.1  &   42.0       &  19.1   \\ \hline
WAN    &   64.8  &   \underline{46.3}       & \underline{20.3}    \\ \hline
ROLE   &    \underline{66.3}  &  43.1        &    15.0 \\ \hline
Ours   &  \textbf{72.0}    & \textbf{48.3}         & \textbf{24.7}    \\ \hline
\end{tabular}
\label{table.spl}
\end{center}
\end{table}

\begin{figure}[ht]
  \centering
   \includegraphics[width=0.45\textwidth]{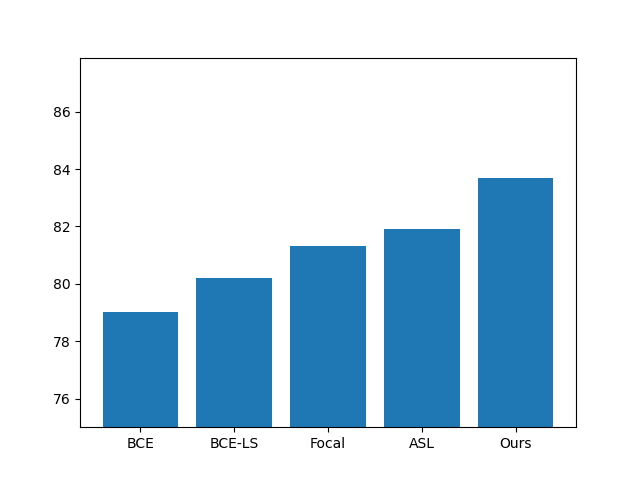}
    \caption{The mAP scores on Open Images dataset. We select BCE, BCE-LS, FOCAL, and ASL for comparison to prove the effectiveness of our approach.}
	\label{fig:oi}
\end{figure}

\textbf{Open Images.}
Because the Open Images dataset has only partial labels, we do not perform pre-processing but run our method directly.
The results are shown in Fig.~\ref{fig:oi}.
Our method achieves the highest mAP score, and compared to BCE, our method improves by 4 approximately.

\subsection{Ablation study}
Finally, to analyze the importance of each contribution, we perform an ablation study on COCO and NUS-WIDE datasets in the PAL\_06 setting and PAL\_06 setting.
The experimental results are shown in Table~\ref{table:ablation}.
We remove $\alpha$ and $\beta$, which are described in~\eqref{bcewithpen}, $d$ in~\eqref{d}, $p$ in ~\eqref{penterm}, and $q$ in~\eqref{dynamicloss}, respectively. 
The results show that every component in our method plays a positive role.

\begin{table}[htbp]
\begin{center}
\caption{The ablation study of our method. The results are shown in mAP metrics.}
\begin{tabular}{|c|cc|cc|}
\hline
\multirow{2}{*}{} & \multicolumn{2}{c|}{COCO}              & \multicolumn{2}{c|}{NUS-WIDE}          \\ \cline{2-5} 
                  & \multicolumn{1}{c|}{PAL\_06} & PPL\_06 & \multicolumn{1}{c|}{PAL\_06} & PPL\_06 \\ \hline
Our method        & \multicolumn{1}{c|}{76.3}        &   80.1      & \multicolumn{1}{c|}{49.7}        & 54.7        \\ \hline
remove  $\alpha$ and $\beta$ Equ.~\ref{bcewithpen}          & \multicolumn{1}{c|}{75.8}        &   79.3      & \multicolumn{1}{c|}{47.6}        &    50.0    \\ \hline
remove  $d$ Equ.~\ref{d}          & \multicolumn{1}{c|}{75.3}        &    78.6     & \multicolumn{1}{c|}{48.0}        &    50.6     \\ \hline
remove   $p$ Equ.~\ref{penterm}         & \multicolumn{1}{c|}{74.9}        & 79.0        & \multicolumn{1}{c|}{45.4}        &   49.0      \\ \hline
remove   $q$ Equ.\ref{dynamicloss}       & \multicolumn{1}{c|}{74.3}        &  78.9       & \multicolumn{1}{c|}{45.6}        & 48.9        \\ \hline
\end{tabular}
\label{table:ablation}
\end{center}
\end{table}

\section{Conclusion}
\label{sec:con}
In this work, we investigate the problem of training multi-label classifiers using partially labeled data to significantly reduce manual labeling costs while maintaining model accuracy. We specifically address the issue of label imbalance, which has long been a challenge in multi-classification problems and is particularly pronounced in the context of partially labeled data.
To address this issue, we first introduce the pseudo-labeling technique, which enables the conventional multi-label model to be directly applied in partial labeling settings without requiring additional structures.
Then, by designing a novel loss function and presenting a dynamically changing training scheme, we effectively alleviate the label imbalance, even in large-scale label spaces.
Lastly, extensive experiments on four commonly used image datasets including COCO, NUS-WIDE, CUB, and Open Images, demonstrate the effectiveness of our proposed approach.

\bibliographystyle{IEEEtran}
\bibliography{reference}

\end{document}